\documentclass[pdflatex,sn-mathphys-num]{sn-jnl}

\usepackage{tabularx}
\usepackage{graphicx}%
\usepackage{multirow}%
\usepackage{amsmath,amssymb,amsfonts}%
\usepackage{amsthm}%
\usepackage{mathrsfs}%
\usepackage[title]{appendix}%
\usepackage{xcolor}%
\usepackage{textcomp}%
\usepackage{manyfoot}%
\usepackage{booktabs}%
\usepackage{algorithm}%
\usepackage{algorithmicx}%
\usepackage{algpseudocode}%
\usepackage{listings}%
\renewcommand{\arraystretch}{1.15}

\theoremstyle{thmstyleone}%
\theoremstyle{thmstyletwo}%

\theoremstyle{thmstylethree}%

\begin{document}

\title[Article Title]{Interpretable Fuzzy Inference for UAV Target Tracking Using Bounding-Box Geometry}



\author[1]{\fnm{Reza} \sur{Ahmari}}\email{rahmari@aggies.ncat.edu}

\author[2]{\fnm{Ahmad} \sur{Mohammadi}}\email{amohammadi@aggies.ncat.edu}
\author[2]{\fnm{Dr. Vahid} \sur{Hemmati}}\email{vhemmati@ncat.edu}
\author[2]{\fnm{Nicholas} \sur{Edmond}}\email{nedmond@aggies.ncat.edu}
\author[2]{\fnm{Hossein} \sur{Z. Saghazadeh}}\email{hzamanisaghazadeh@aggies.ncat.edu}

\author[1]{\fnm{Dr. Olusola} \sur{Odeyomi}}\email{otodeyomi@ncat.edu}
\author[2]{\fnm{Dr. Parham} \sur{Kebria}}\email{pmkebria@ncat.edu}

\author*[2]{\fnm{Dr. Abdollah} \sur{Homaifar}}\email{homaifar@ncat.edu}

\affil*[1]{\orgdiv{Department of Computer Science}, \orgname{North Carolina A\&T State University}, \orgaddress{\street{1601 E Market St}, \city{Greensboro}, \postcode{27411}, \state{NC}, \country{USA}}}

\affil[2]{\orgdiv{Department of Electrical and Computer Engineering}, \orgname{North Carolina A\&T State University}, \orgaddress{\street{1601 E Market St}, \city{Greensboro}, \postcode{27411}, \state{NC}, \country{USA}}}




\abstract{Vision-based guidance of unmanned aerial vehicles (UAVs) toward unmanned ground vehicles (UGVs) is a key capability for cooperative aerial--ground robotic systems; however, estimating a reliable continuous yaw command from onboard visual perception alone remains challenging due to sensing uncertainty, limited onboard computational resources, and the need for interpretable control strategies. Many existing approaches rely on deep learning or geometric reconstruction methods that require large datasets, external localization, or complex modeling assumptions, reducing transparency and suitability for real-time deployment on resource-constrained platforms. 
This paper presents an interpretable fuzzy-inference-based framework for generating continuous yaw commands using only low-dimensional vision features extracted from YOLO bounding boxes, including target centroid location, area, and aspect ratio, without explicit geometric modeling. A Mamdani fuzzy system is developed as an interpretable baseline using a shoulder--triangle--shoulder input partition, followed by a first-order Takagi--Sugeno model with three antecedent membership terms per input, whose parameters are derived from training-set quantiles, yielding a compact 27-rule structure.
The approach is evaluated on 6{,}169 labeled samples collected in a VICON motion-capture environment. Across five randomized train--test splits, the Takagi--Sugeno model achieves a test-set mean absolute error of $0.140^\circ \pm 0.003^\circ$, a root mean squared error of $0.200^\circ \pm 0.008^\circ$, and a maximum absolute error of $1.254^\circ \pm 0.121^\circ$. The within-threshold accuracies are $99.676\% \pm 0.270\%$ for $\pm1^\circ$ and $100.000\% \pm 0.000\%$ for both $\pm3^\circ$ and $\pm5^\circ$. Directional consistency between image-plane horizontal displacement and predicted yaw sign reaches $90.254\% \pm 0.612\%$, demonstrating that the proposed framework provides a transparent, data-efficient, and computationally lightweight solution for real-time vision-based UAV guidance toward mobile ground targets.}




\maketitle

\section{Introduction}\label{sec1}

Vision-based perception plays a central role in autonomous aerial and ground robotic systems, particularly in scenarios where precise relative positioning and alignment are required for navigation, tracking, or cooperative tasks \cite{hutchinson2002tutorial,chaumette2006visual}. 
Unmanned aerial vehicles (UAVs) are increasingly deployed across a wide range of applications including surveillance, infrastructure inspection, logistics, and cooperative aerial–ground robotics, highlighting the growing importance of reliable onboard perception and navigation capabilities \cite{mohsan2023unmanned}. In such settings, compact geometric descriptors derived from object detections, such as bounding-box location, scale, and shape, provide an attractive alternative to dense image representations due to their low dimensionality and robustness to appearance variations.

A recurring challenge in vision-to-control pipelines is the reliable mapping from noisy, perspective-dependent visual features to continuous control commands.
Recent data-driven approaches, including deep neural networks and attention-based models, have demonstrated strong representational power for vision-based perception and regression tasks \cite{bojarski2016end,muhammad2020deep,liu2024explainable,ming2024not}. While such methods can achieve high accuracy, they typically require large annotated datasets, introduce a high number of learned parameters, and rely on complex feature interactions that are difficult to interpret and validate within safety-critical control loops. In addition, their performance can be sensitive to distributional shifts in visual appearance, scale, or viewpoint, which are common in aerial robotics scenarios. In contrast, rule-based fuzzy inference systems explicitly encode the mapping between visual descriptors and control actions through transparent linguistic rules, enabling predictable behavior, easier validation, and graceful degradation under uncertainty \cite{mamdani1975experiment, passino1998fuzzy}. These properties make fuzzy logic particularly well suited for low-dimensional, structured inputs such as bounding-box geometry, where interpretability, robustness, and controlled generalization are often more critical than maximizing perceptual expressiveness.

Classical Mamdani fuzzy systems are widely used in control applications due to their transparency and intuitive linguistic structure.
However, when applied to continuous regression tasks, Mamdani inference is known to exhibit limited numerical precision and output saturation effects, particularly as the dimensionality of the input space increases \cite{passino1998fuzzy}.
Takagi--Sugeno fuzzy systems address these limitations by replacing fuzzy consequents with rule-local linear models, yielding improved approximation capability while preserving the interpretability of fuzzy antecedents \cite{takagi1985fuzzy, sugeno1985industrial}.

This paper presents a systematic and reproducible fuzzy-logic framework for estimating a continuous yaw correction command from YOLO bounding-box descriptors \cite{redmon2016you}. The objective is not to introduce a new fuzzy-inference family or a new visual descriptor, but to formulate and validate a leakage-controlled, low-dimensional vision-to-yaw mapping in which interpretable fuzzy reasoning is applied directly to detector-produced visual perception outputs.

Two fuzzy inference formulations are evaluated under a shared three-input antecedent rule structure. First, a Mamdani fuzzy controller is constructed as an interpretable baseline using robust percentile-based normalization, explicit shoulder--triangle--shoulder membership functions, and a deterministic same-side rule design. Second, a first-order Takagi--Sugeno model is constructed using train-derived quantile membership parameters and rule-local linear consequents identified by least-squares regression. In both cases, the antecedent space is formed from the same three YOLO-derived bounding-box descriptors: horizontal center, normalized area, and aspect ratio.

The technical contribution lies in the controlled formulation of a compact fuzzy yaw-estimation pipeline that bridges detector-based visual perception and fuzzy inference. The framework combines: (i) training-only feature and membership-function parameterization to avoid test-data leakage, (ii) coverage-safe shoulder--triangle--shoulder antecedent partitions to reduce boundary activation ambiguity, (iii) a shared interpretable antecedent rule structure for Mamdani and Takagi--Sugeno inference, and (iv) multi-metric evaluation against VICON-derived yaw labels and standard low-dimensional regression baselines. This design clarifies the trade-off between transparent fuzzy directional reasoning and high-precision continuous yaw regression.

The contributions supported by this paper are as follows:
\begin{enumerate}
    \item A reproducible low-dimensional vision-to-yaw estimation framework that maps YOLO bounding-box geometry to a continuous yaw correction variable using only horizontal center, normalized area, and aspect ratio.
    
    \item A leakage-controlled membership-function design in which feature normalization and fuzzy antecedent parameters are derived from the training split only and applied unchanged during test-time inference.
    
    \item A coverage-safe shoulder--triangle--shoulder fuzzy partition for the antecedent variables, reducing boundary activation ambiguity while preserving a compact 27-rule fuzzy structure.
    
    \item A controlled comparison between Mamdani and first-order Takagi--Sugeno inference under a shared antecedent rule structure, clarifying the trade-off between interpretable directional reasoning and high-precision continuous yaw estimation.
    
    \item A comparative evaluation against standard low-dimensional regression baselines and computational-volume measurements, enabling assessment of accuracy, worst-case error, side-consistency, interpretability, and inference cost.
\end{enumerate}


\section{Related Work}\label{sec2}

Vision-based heading and yaw estimation has been studied extensively in the contexts of autonomous navigation, visual servoing, and cooperative aerial--ground systems. Existing approaches can be broadly categorized into geometry-based methods, data-driven regression models, and fuzzy or rule-based controllers \cite{hutchinson2002tutorial,chaumette2006visual,wu2022survey}. This section reviews prior work along these axes, with emphasis on methods that leverage bounding-box or image-plane features and those that balance interpretability with predictive accuracy.

\subsection{Vision-Based Heading and Yaw Estimation}\label{subsec1} Early work on vision-based yaw and heading estimation largely relied on explicit geometric reasoning and visual servoing principles. Classical image-based visual servoing (IBVS) formulations estimate orientation errors directly from image features such as centroids, edges, or point correspondences, and generate control commands through analytical Jacobians \cite{hutchinson2002tutorial, chaumette2006visual}. While these methods provide theoretical guarantees under idealized conditions, their performance degrades in the presence of detection noise, partial occlusion, and imperfect feature association, which are common in real-world robotic deployments.

More recent studies have leveraged object detection outputs, particularly bounding boxes, as compact representations of target pose in the image plane. Bounding-box center displacement provides a direct image-plane cue for lateral misalignment, while box area and scale variation act as depth-related cues in monocular settings through perspective projection and region-feature geometry \cite{hutchinson2002tutorial,chaumette2006visual,chaumette2004image}. These approaches reduce reliance on precise feature tracking and enable integration with modern deep detectors, but they require robust mappings from low-dimensional visual descriptors to continuous control signals \cite{wu2022survey}.

\subsection{Data-Driven Regression Approaches}\label{subsec2}
 With the increasing availability of labeled data, fully data-driven models have been proposed to regress yaw or steering commands directly from visual inputs or intermediate features. Convolutional neural networks have been used to predict steering angles or yaw corrections end-to-end from raw images, as demonstrated in learning-based driving and aerial navigation systems \cite{bojarski2016end}. Although such models can achieve high numerical accuracy, they typically require large training datasets, exhibit limited interpretability, and provide little insight into failure modes when operating outside the training distribution \cite{muhammad2020deep}.

Hybrid approaches have also been explored, where deep networks extract intermediate features that are subsequently mapped to control variables using shallow regressors or linear models. While these methods reduce model complexity relative to end-to-end learning, they still rely on opaque feature representations and often lack explicit mechanisms for incorporating domain knowledge or enforcing monotonic relationships between visual cues and control outputs.

\subsection{Fuzzy Logic Controllers in Robotics}\label{subsec3} Fuzzy logic control has a long history in mobile robotics and autonomous systems, particularly in scenarios involving uncertainty, imprecise sensing, or heuristic reasoning. Mamdani-type fuzzy controllers have been widely applied to navigation, obstacle avoidance, and alignment tasks due to their linguistic rule structure and intuitive interpretability \cite{mamdani1975experiment, passino1998fuzzy}. In visual navigation contexts, fuzzy rules such as ``if target is left, turn left'' provide transparent decision logic that can be inspected and modified by human designers.

Prior work has shown that Mamdani-type fuzzy inference systems, while highly interpretable and well suited for linguistic control design, can exhibit limitations when applied to continuous regression and estimation problems. Because the output of a Mamdani system is synthesized through fixed output membership functions and defuzzification, the resulting mapping may suffer from limited numerical resolution and sensitivity to the chosen output partition, particularly when modeling smooth input--output relationships. Such characteristics can lead to coarse control signals and reduced approximation accuracy in strongly continuous domains \cite{mendel2002type,riza2015frbs}. In contrast, fuzzy rule-based systems with functional consequents, including Takagi--Sugeno formulations, are explicitly designed to improve regression performance by representing local input--output relationships more precisely.

Takagi--Sugeno fuzzy systems have been proposed as a remedy to these limitations by replacing fuzzy-set consequents with parametric functions, typically linear in the inputs. In robotics applications, first-order Sugeno models have demonstrated improved approximation capability while retaining the structured partitioning of the input space provided by fuzzy membership functions \cite{takagi1985fuzzy, sugeno1985industrial}. More broadly, robust yaw regulation remains an important control problem across autonomous robotic platforms. For example, Karade et al.~\cite{karade2025robustyaw} investigated robust yaw angle control for autonomous underwater vehicles using a dynamic surface-based optimized second-order sliding-mode framework under disturbances and model uncertainty. Although that work focuses on model-based control rather than vision-based fuzzy inference, it further highlights the importance of accurate and robust yaw-related decision making in autonomous robotic systems. The consequent parameters can be identified using least-squares techniques, enabling data-driven optimization without sacrificing the interpretability of the rule antecedents.
Hybrid fuzzy–neural approaches have also been explored to enhance robustness and adaptability in uncertain control environments \cite{kebria2019type, kebria2019adaptive}.

\subsection{Bounding-Box Features and Normalization}\label{subsec4} Several studies have emphasized the importance of feature normalization when using bounding-box descriptors for control or estimation. Bounding-box area and aspect ratio are particularly sensitive to distance changes, partial visibility, and detector noise. Percentile-based or quantile-based normalization schemes have been employed in related contexts to mitigate the influence of outliers and to stabilize downstream learning algorithms \cite{hastie2009elements, kebria2019fuzzy, loh2025theoretical}.

Despite their practical importance, preprocessing and membership-function parameterization strategies are often under-specified in fuzzy perception pipelines, leading to ambiguity regarding training--test leakage and reproducibility. In the present framework, all feature-scaling and membership-function parameters are derived from the training split only and applied unchanged during test-time inference. This ensures that the reported Mamdani and Takagi--Sugeno results are not affected by test-set information leakage.

\subsection{Positioning of the Present Work}\label{subsec5} 
The primary contribution of this work is therefore not a new fuzzy-inference formulation itself, but a framework-level integration in which interpretable fuzzy reasoning is applied directly to detector-produced visual perception outputs rather than to direct physical state measurements. Specifically, YOLO-derived bounding-box descriptors are used as compact geometric inputs for monocular UAV--UGV relative yaw estimation, establishing a lightweight perception-level fuzzy inference framework that bridges object-detection-based visual perception and fuzzy reasoning.

Unlike prior fuzzy-control studies that emphasize heuristic tuning or simulation-only validation, the proposed approach is evaluated on a real-world dataset with high-fidelity VICON-based ground-truth yaw measurements. By evaluating Mamdani inference, Takagi--Sugeno inference, standard regression baselines, and computational cost under the same bounding-box feature setting, the study clarifies the trade-offs among interpretability, numerical accuracy, worst-case error, side-consistency, and inference efficiency in vision-based UAV yaw estimation.


\section{Methodology}\label{sec3}
The objective of this section is to develop a reproducible pipeline for predicting a continuous yaw correction command, $\theta$ (degrees), from vision-based features.

Each data sample corresponds to a single image frame containing a YOLO detected UGV bounding box, from which geometric descriptors are extracted. The corresponding ground-truth yaw reference is obtained from a VICON motion-capture system.
Two fuzzy inference systems are evaluated using an identical antecedent structure with three input variables and three linguistic terms per input (27 rules total). First,  a Mamdani baseline designed for interpretability, and secondly,   a first-order Takagi--Sugeno model, in which the consequent parameters are identified by least squares on the training split.

\subsection{Notation and Problem Definition}\label{subsec1}
Each sample corresponds to one image frame and one detected target bounding box. Let the image resolution be $W\times H$ pixels and the detected bounding box be
\begin{equation}
\begin{aligned}
\mathcal{B}=(x_1,y_1,x_2,y_2), \\ \qquad
0 \le x_1 < x_2 \le W,\;\; 0 \le y_1 < y_2 \le H.
\end{aligned}
\end{equation}
where $(x_1,y_1)$, and $(x_2,y_2)$  denote the pixel coordinates of the top-left and bottom-right corners of the bounding box, respectively.

The goal is to estimate a yaw $\hat{\theta}\in\mathbb{R}$ from a three-dimensional feature vector
\begin{equation}
\mathbf{x}=[c_x,\; a,\; r]^{\mathsf{T}}\in\mathbb{R}^3
\end{equation}
via a mapping $\hat{\theta}=f(\mathbf{x})$. Where, $( c_x,\; a,\; r )$, denote the horizontal center, normalized area, and aspect ratio features extracted from the bounding box. Depending on the model formulation, these features may be used directly or after normalization. These parameters are defined in the following section.

In this paper we evaluate the perception-to-yaw mapping offline to isolate inference accuracy and
generalization. In deployment, $\hat{\theta}$ is intended to serve as a yaw correction reference that can
be fed to a low-level yaw-rate or attitude controller within the UAV’s onboard control loop.
\subsection{Ground-Truth Yaw from VICON}\label{subsec2}
Ground-truth yaw labels are generated from a VICON motion-capture system providing millimeter-level positional accuracy \cite{merriaux2017study}.
Let $\mathbf{p}_u=[p_{u,x},p_{u,y},p_{u,z}]^{\mathsf{T}}$ and $\mathbf{p}_g=[p_{g,x},p_{g,y},p_{g,z}]^{\mathsf{T}}$
denote the UAV and UGV positions in the VICON frame.
The horizontal relative position of UGV with respect to the position of UAV is
\begin{equation}
\label{eq:v1}
\Delta \mathbf{p}_{xy}=
\begin{bmatrix}
p_{g,x}-p_{u,x}\\
p_{g,y}-p_{u,y}
\end{bmatrix}.
\end{equation}
The ground-truth yaw label $\theta$ is defined as the signed line-of-sight bearing from the UAV to the UGV:
\begin{equation}
\label{eq:v2}
\theta = \operatorname{atan2}\!\left(p_{g,y}-p_{u,y},\;p_{g,x}-p_{u,x}\right),
\end{equation}
For reporting and evaluation, $\theta$ is expressed in degrees.
Negative values correspond to leftward bearings and positive values correspond to rightward bearings under the adopted coordinate convention.

\subsection{Feature Definitions from Bounding Boxes}\label{subsec3}
The dataset includes the raw bounding-box coordinates $(x_1,y_1,x_2,y_2)$ and the derived features $(c_x,c_y,a,r)$.
We use $c_x$, $a$, and $r$ as inputs. For completeness, the feature definitions are listed below.

\paragraph{Normalized Horizontal Center}\label{subsubsec1}
The bounding-box center is
\begin{equation}
x_c=\frac{x_1+x_2}{2},\qquad y_c=\frac{y_1+y_2}{2},
\end{equation}
and the normalized horizontal center is
\begin{equation}
c_x=\frac{x_c}{W}\in[0,1].
\end{equation}
Although $c_y=y_c/H$ is available, it is not used in the three-input models reported here.

\paragraph{Normalized Area}\label{subsubsec2}
Let $w=x_2-x_1$ and $h=y_2-y_1$ denote the bounding-box width and height (pixels).
The normalized area is
\begin{equation}
a=\frac{wh}{WH}\in[0,1].
\end{equation}

\paragraph{Aspect Ratio}\label{subsubsec3}
In our experiments, bounding-box aspect ratio $r$ is used as the third input to both fuzzy models and matches the dataset column \texttt{aspect\_ratio}.
It is defined as
\begin{equation}
r=\frac{h}{w},
\end{equation}
With this definition, larger $r$ corresponds to a taller or narrower box, while smaller $r$ corresponds to a wider box. This convention yields a positive feature that captures perspective-induced shape distortion and remains numerically stable under aerial viewpoints.

\vspace{1em}
Let $\mathcal{D}=\{(\mathbf{x}_n,\theta_n)\}_{n=1}^{N}$ with $N=6169$.
Indices are shuffled and split into training and test sets. To account for split variability, all experiments are repeated over five randomized splits with different fixed seeds:
\begin{equation}
\mathcal{I}_{\mathrm{train}}\cup\mathcal{I}_{\mathrm{test}}=\{1,\ldots,N\},\qquad
\mathcal{I}_{\mathrm{train}}\cap\mathcal{I}_{\mathrm{test}}=\emptyset,
\end{equation}
with $|\mathcal{I}_{\mathrm{train}}|=\lfloor0.7N\rfloor$ and $|\mathcal{I}_{\mathrm{test}}|=N-|\mathcal{I}_{\mathrm{train}}|$.

\vspace{1em}

To reduce sensitivity to outliers and to stabilize membership-function design, the Mamdani baseline uses robust percentile-based normalization.
For each feature $x\in\{c_x,a,r\}$, compute on the training set only:
\begin{equation}
\begin{aligned}
q_{0.05}(x)=Q_{0.05}(\{x_n:n\in\mathcal{I}_{\mathrm{train}}\}), \\
q_{0.95}(x)=Q_{0.95}(\{x_n:n\in\mathcal{I}_{\mathrm{train}}\}),
\end{aligned}
\end{equation}
where $Q_p(\cdot)$ denotes the empirical $p$-quantile.
Each sample (train or test) is then clipped and mapped to $[0,1]$:
\begin{align}
x_{\mathrm{clip}} &= \min\big(\max(x,\;q_{0.05}(x)),\;q_{0.95}(x)\big),\\
x_{\mathrm{norm}} &= \frac{x_{\mathrm{clip}}-q_{0.05}(x)}{q_{0.95}(x)-q_{0.05}(x)+\varepsilon}\in[0,1].
\end{align}
This yields $\mathbf{x}_{\mathrm{norm}}=[c_{x,\mathrm{norm}},a_{\mathrm{norm}},r_{\mathrm{norm}}]^{\mathsf{T}}$ for the Mamdani controller.
For the Takagi--Sugeno model, membership parameters are derived from training quantiles directly in the original feature scale (Section~\ref{sec:sugeno_mf}), and no additional robust min--max normalization is applied.

\vspace{1em}

The fuzzy partitions used in this work combine triangular and shoulder membership functions. Central linguistic terms are represented using triangular membership functions, while boundary terms are represented using left- and right-shoulder functions to ensure nonzero coverage at the extremes. For a central triangular term with parameters $(\alpha,\beta,\gamma)$, where $\alpha\le\beta\le\gamma$, we define
\begin{equation}
\mu_{\mathrm{tri}}(x;\alpha,\beta,\gamma)=
\max\!\left(0,\;
\min\!\left(\frac{x-\alpha}{\beta-\alpha+\varepsilon},\;
\frac{\gamma-x}{\gamma-\beta+\varepsilon}\right)\right),
\end{equation}
which is bounded in $[0,1]$ and numerically stable due to $\varepsilon>0$.

\subsection{Mamdani Fuzzy Inference System}\label{subsec4}
The Mamdani controller operates on the normalized inputs $\mathbf{x}_{\mathrm{norm}}$ and assigns three linguistic terms to each input \cite{mamdani1975experiment, passino1998fuzzy}.
For the lateral position feature $c_x$, the terms are \emph{Left}, \emph{Center}, and \emph{Right}.
For the area feature $a$ (a distance proxy, since larger image area indicates a closer target), the terms are \emph{Far}, \emph{Mid}, and \emph{Near}.
For the aspect ratio feature $r$, the terms are \emph{Wide}, \emph{Normal}, and \emph{Tall}.

All terms are implemented using the same three-term shoulder--triangle--shoulder partition on $[0,1]$; only the linguistic names differ by input. To ensure nonzero membership at the endpoints, the \emph{Left} and \emph{Right} terms are implemented as shoulder functions, while the \emph{Center} term is implemented as a triangular function:
\begin{equation}
\begin{gathered}
\mu_{L}(x)=
\begin{cases}
1, & x \le 0,\\[3pt]
\dfrac{0.5-x}{0.5+\varepsilon}, & 0 < x < 0.5,\\[6pt]
0, & x \ge 0.5,
\end{cases}
\qquad
\mu_{C}(x)=\mu_{\mathrm{tri}}(x;0,0.5,1),
\qquad\\
\mu_{R}(x)=
\begin{cases}
0, & x \le 0.5,\\[3pt]
\dfrac{x-0.5}{0.5+\varepsilon}, & 0.5 < x < 1,\\[6pt]
1, & x \ge 1.
\end{cases}
\end{gathered}
\end{equation}

When applied to $a_{\mathrm{norm}}$, the sets $\{\mu_L,\mu_C,\mu_R\}$ correspond to \{\emph{Far}, \emph{Mid}, \emph{Near}\}; when applied to $r_{\mathrm{norm}}$, they correspond to \{\emph{Wide}, \emph{Normal}, \emph{Tall}\}.

With three inputs and three terms per input, the rule base contains $N_r=3^3=27$ rules, each rule i has the form:
\begin{equation}
\mathcal{R}_i:\ \text{If } c_x \text{ is } A_i \text{ and } a \text{ is } B_i \text{ and } r \text{ is } C_i,\ \text{then } \theta \text{ is } D_i,
\end{equation}
where $A_i,B_i,C_i\in\{L,C,R\}$ and $D_i$ is one of five output labels.

\paragraph{Rule Firing Strength}\label{subsubsec1}
Given $\mathbf{x}_{\mathrm{norm}}$, rule $i$ fires with strength (minimum t-norm)
\begin{equation}
w_i=\min\!\Big(\mu_{A_i}(c_{x,\mathrm{norm}}),\;\mu_{B_i}(a_{\mathrm{norm}}),\;\mu_{C_i}(r_{\mathrm{norm}})\Big).
\end{equation}

\paragraph{Output Sets and Defuzzification}\label{subsubsec2}
Let $\theta_{\min}$ and $\theta_{\max}$ denote the minimum and maximum training labels, respectively.
Five triangular output membership functions are constructed from $[\theta_{\min},\theta_{\max}]$ using a padding
$\mathrm{pad}_\theta=0.05(\theta_{\max}-\theta_{\min})$ and uniform breakpoints over
$[\ell,h]=[\theta_{\min}-\mathrm{pad}_\theta,\ \theta_{\max}+\mathrm{pad}_\theta]$:

\begin{equation}
\begin{aligned}
\mu_{\text{SharpLeft}}(\theta)  &= \mu_{\mathrm{tri}}(\theta;\ell,\ell,\ell+0.25(h-\ell)),\\
\mu_{\text{Left}}(\theta)       &= \mu_{\mathrm{tri}}(\theta;\ell,\ell+0.25(h-\ell),\ell+0.50(h-\ell)),\\
\mu_{\text{Zero}}(\theta)       &= \mu_{\mathrm{tri}}(\theta;\ell+0.25(h-\ell),\ell+0.50(h-\ell),\ell+0.75(h-\ell)),\\
\mu_{\text{Right}}(\theta)      &= \mu_{\mathrm{tri}}(\theta;\ell+0.50(h-\ell),\ell+0.75(h-\ell),h),\\
\mu_{\text{SharpRight}}(\theta) &= \mu_{\mathrm{tri}}(\theta;\ell+0.75(h-\ell),h,h).
\end{aligned}
\end{equation}

Mamdani implication uses clipping:
\begin{equation}
\mu_i(\theta)=\min\big(w_i,\mu_{D_i}(\theta)\big),
\end{equation}
and aggregation uses the maximum operator:
\begin{equation}
\mu_{\mathrm{out}}(\theta)=\max_{i=1,\ldots,27}\mu_i(\theta).
\end{equation}
The crisp output is computed by centroid defuzzification over
$\Theta=[\theta_{\min}-5^\circ,\theta_{\max}+5^\circ]$:
\begin{equation}
\hat{\theta}=
\frac{\int_{\Theta}\theta\,\mu_{\mathrm{out}}(\theta)\,d\theta}
{\int_{\Theta}\mu_{\mathrm{out}}(\theta)\,d\theta}.
\end{equation}
In implementation, the integrals are approximated on a uniform grid with $M=1201$ points.

\subsubsection{Same-Side Rule Logic}\label{subsubsec3}

The rule generator is designed to promote sign consistency between the lateral displacement of the target in the image plane and the predicted yaw command.
The normalized horizontal center feature $c_x$ determines the turn direction, where \emph{Left} corresponds to a negative yaw correction, \emph{Right} corresponds to a positive yaw correction, and \emph{Center} corresponds to a near-zero yaw command.

The normalized area feature $a_{\mathrm{norm}}$ and the aspect ratio feature
$r_{\mathrm{norm}}$ modulate the aggressiveness of the yaw response.
As an example, turns are sharpened when the target is categorized as
\emph{Near} (large $a_{\mathrm{norm}}$), indicating close proximity, and/or
\emph{Tall} (large $r_{\mathrm{norm}}$), corresponding to vertically elongated
or narrow bounding boxes.
These conditions empirically correlate with increased sensitivity to lateral
misalignment under the adopted aerial viewing geometry.

This same-side mapping encourages the direction of the yaw command to remain consistent with the visual lateral offset of the target, while allowing the magnitude of the response to adapt smoothly based on target scale and shape.

\paragraph{Rule Base Specification}
The complete 27-rule base for both the Mamdani and Takagi--Sugeno systems is deterministically generated from the Cartesian product of the three linguistic terms for each input:
$\{\text{Left},\text{Center},\text{Right}\}$ for $c_x$,
$\{\text{Far},\text{Mid},\text{Near}\}$ for $a$,
and $\{\text{Wide},\text{Normal},\text{Tall}\}$ for $r$.
In the Mamdani controller, each antecedent combination is assigned a linguistic consequent label $D_i$ according to the same-side mapping described above.
In the Takagi--Sugeno model, the same antecedent grid is retained, but each rule uses a learned first-order consequent
$z_i(\mathbf{x})=p_i c_x+q_i a+s_i r+t_i$
instead of a fixed linguistic consequent label.
Because the antecedent grid is generated algorithmically and contains 27 rules, the full rule table is provided in Appendix~\ref{tab:rulebase27} for completeness and reproducibility.

\subsection{First-Order Takagi--Sugeno Fuzzy Inference System}\label{subsec1}
The Takagi--Sugeno model uses the same 27-rule antecedent structure but replaces fuzzy consequents with rule-local linear functions \cite{takagi1985fuzzy, sugeno1985industrial}.
For readability, we denote the three fuzzy sets for $c_x$ as \emph{Left}, \emph{Center}, \emph{Right}; for $a$ as \emph{Far}, \emph{Mid}, \emph{Near}; and for $r$ as \emph{Wide}, \emph{Normal}, \emph{Tall}, although the underlying three-term parameterization follows the same left-shoulder, center-triangle, right-shoulder construction in each case. Unless otherwise stated, this model operates on the original feature scale $(c_x,a,r)$.

\paragraph{Quantile-Derived Membership Parameters (Coverage-Guaranteed).}
\label{sec:sugeno_mf}
For each input variable $x\in\{c_x,a,r\}$, training-set quantiles
$q_{0.05}(x)$, $q_{0.50}(x)$, and $q_{0.95}(x)$ are computed using only samples in $\mathcal{I}_{\mathrm{train}}$.
To guarantee nonzero fuzzy coverage and prevent cases where all rule firing strengths become zero, the Takagi--Sugeno model uses a three-term partition consisting of a left shoulder function, a central triangular function, and a right shoulder function:
\begin{align}
\mu_{\text{Left}}(x) &=
\begin{cases}
1, & x \le q_{0.05}(x),\\[3pt]
\dfrac{q_{0.50}(x)-x}{q_{0.50}(x)-q_{0.05}(x)+\varepsilon}, & q_{0.05}(x) < x < q_{0.50}(x),\\[6pt]
0, & x \ge q_{0.50}(x),
\end{cases}\\
\mu_{\text{Center}}(x) &= \mu_{\mathrm{tri}}\!\big(x; q_{0.05}(x),\,q_{0.50}(x),\,q_{0.95}(x)\big),\\
\mu_{\text{Right}}(x) &=
\begin{cases}
0, & x \le q_{0.50}(x),\\[3pt]
\dfrac{x-q_{0.50}(x)}{q_{0.95}(x)-q_{0.50}(x)+\varepsilon}, & q_{0.50}(x) < x < q_{0.95}(x),\\[6pt]
1, & x \ge q_{0.95}(x).
\end{cases}
\end{align}
All quantiles are computed on $\mathcal{I}_{\mathrm{train}}$ only and held fixed for test-time inference.

\paragraph{Firing Strengths, Normalization, and Output}\label{subsubsec1}
For each rule $i$, the firing strength is computed using the \emph{product} t-norm:
\begin{equation}
w_i=\mu_{A_i}(c_x)\,\mu_{B_i}(a)\,\mu_{C_i}(r),
\end{equation}
and normalized weights are
\begin{equation}
\bar{w}_i=\frac{w_i}{\sum_{k=1}^{27} w_k+\varepsilon}.
\end{equation}
Each rule has a first-order consequent
\begin{equation}
z_i(\mathbf{x})=p_i c_x + q_i a + s_i r + t_i,
\end{equation}
and the Sugeno output is
\begin{equation}
\hat{\theta}=\sum_{i=1}^{27}\bar{w}_i\,z_i(\mathbf{x}).
\end{equation}

\paragraph{Least-Squares Identification}\label{subsubsec1}
Let $\beta\in\mathbb{R}^{108}$ collect all consequent coefficients
$\{p_i,q_i,s_i,t_i\}_{i=1}^{27}$.
For each training sample $n$, construct the regression row
$\boldsymbol{\phi}^{(n)}\in\mathbb{R}^{108}$ by stacking
$\{\bar{w}_i^{(n)}c_x^{(n)},\bar{w}_i^{(n)}a^{(n)},\bar{w}_i^{(n)}r^{(n)},\bar{w}_i^{(n)}\}_{i=1}^{27}$.
Stacking all training rows yields $\Phi\in\mathbb{R}^{N_{\mathrm{train}}\times108}$ and labels $\mathbf{y}\in\mathbb{R}^{N_{\mathrm{train}}}$.
The parameters are estimated by least squares:
\begin{equation}
\hat{\beta}=\arg\min_{\beta}\|\Phi\beta-\mathbf{y}\|_2^2,
\end{equation}
which is computed using numerically stable solvers (SVD/QR).

At test time, the train-derived membership parameters and the learned $\hat{\beta}$ are fixed.
For each test sample, membership degrees, firing strengths, normalized weights, and $\hat{\theta}$ are computed without re-fitting.

\subsection{Evaluation Metrics}\label{evaluation}
Let $e_i=\hat{\theta}_i-\theta_i$ denote the test error for sample $i$.
We report mean absolute error (MAE), root mean square error (RMSE), maximum absolute error (MAXAE), and within-threshold accuracy for $\tau\in\{1^\circ,3^\circ,5^\circ\}$:

\begin{align}
\mathrm{MAE} &= \frac{1}{N_{\mathrm{test}}}\sum_{i=1}^{N_{\mathrm{test}}}|e_i|,\\
\mathrm{RMSE} &= \sqrt{\frac{1}{N_{\mathrm{test}}}\sum_{i=1}^{N_{\mathrm{test}}}e_i^2},\\
\mathrm{MAXAE} &= \max_{i=1,\ldots,N_{\mathrm{test}}}|e_i|,\\
\mathrm{Acc}_{\pm\tau} &=
\frac{1}{N_{\mathrm{test}}}\sum_{i=1}^{N_{\mathrm{test}}}
\mathbb{I}\big(|e_i|\le\tau\big)\times 100\%.
\end{align}

Within-threshold accuracy at tighter tolerances ($\pm1^\circ$ and $\pm3^\circ$) is included to quantify fine-grained yaw estimation performance near zero, which is particularly relevant for stable closed-loop heading regulation and avoidance of control chatter.

\paragraph{Directional (Side) Consistency (SC)}
In addition to numerical error metrics, we quantify whether the predicted yaw command preserves the intended turning \emph{direction} implied by the image-plane lateral displacement. Let $m_{c_x}$ denote the median of $c_x$ computed on the \emph{training set} only. We define a three-class side label from the horizontal target location:
\begin{equation}
s_{c_x}(c_x)=
\begin{cases}
\text{Left}, & c_x < m_{c_x},\\
\text{Right}, & c_x > m_{c_x},\\
\text{Center}, & \text{otherwise}.
\end{cases}
\label{eq:side_cx}
\end{equation}
Similarly, we map the predicted yaw to a three-class directional label using a small deadband $\delta$ (degrees) to avoid counting near-zero corrections as left/right:
\begin{equation}
s_{\theta}(\hat{\theta})=
\begin{cases}
\text{Left}, & \hat{\theta} < -\delta,\\
\text{Center}, & |\hat{\theta}| \le \delta,\\
\text{Right}, & \hat{\theta} > \delta.
\end{cases}
\label{eq:side_theta}
\end{equation}
Directional (Side) consistency (SC) is then computed on the test set as the percentage of samples whose side labels agree:
\begin{equation}
\mathrm{SideCons}=
\frac{1}{N_{\mathrm{test}}}\sum_{i=1}^{N_{\mathrm{test}}}
\mathbb{I}\!\Big(s_{c_x}(c_{x,i}) = s_{\theta}(\hat{\theta}_i)\Big)\times 100\%.
\label{eq:side_consistency}
\end{equation}
In our experiments, $\delta=3^\circ$ and $m_{c_x}$ is computed from the training split only and held fixed for test evaluation. Side-consistency is used solely as an evaluation metric and is not explicitly enforced during fuzzy inference or parameter estimation.


\section{Dataset Collection and Processing}\label{sec4}
The dataset used in this study was collected and fully documented in our prior work \cite{ahmarijournal, ahmariconf}. In this paper, we reuse the same experimental dataset and ground-truth generation procedure to enable a controlled evaluation of interpretable fuzzy inference systems for vision-based yaw estimation. For completeness and reproducibility, we summarize the key aspects of the hardware setup, data acquisition procedure, and labeling pipeline here, while referring the reader to \cite{ahmarijournal} for full implementation details, system diagrams, and extended experimental context.

\subsection{Experimental Setup}\label{subsec1}
\begin{figure}[h]
    \centering
    \includegraphics[width=0.5\textwidth]{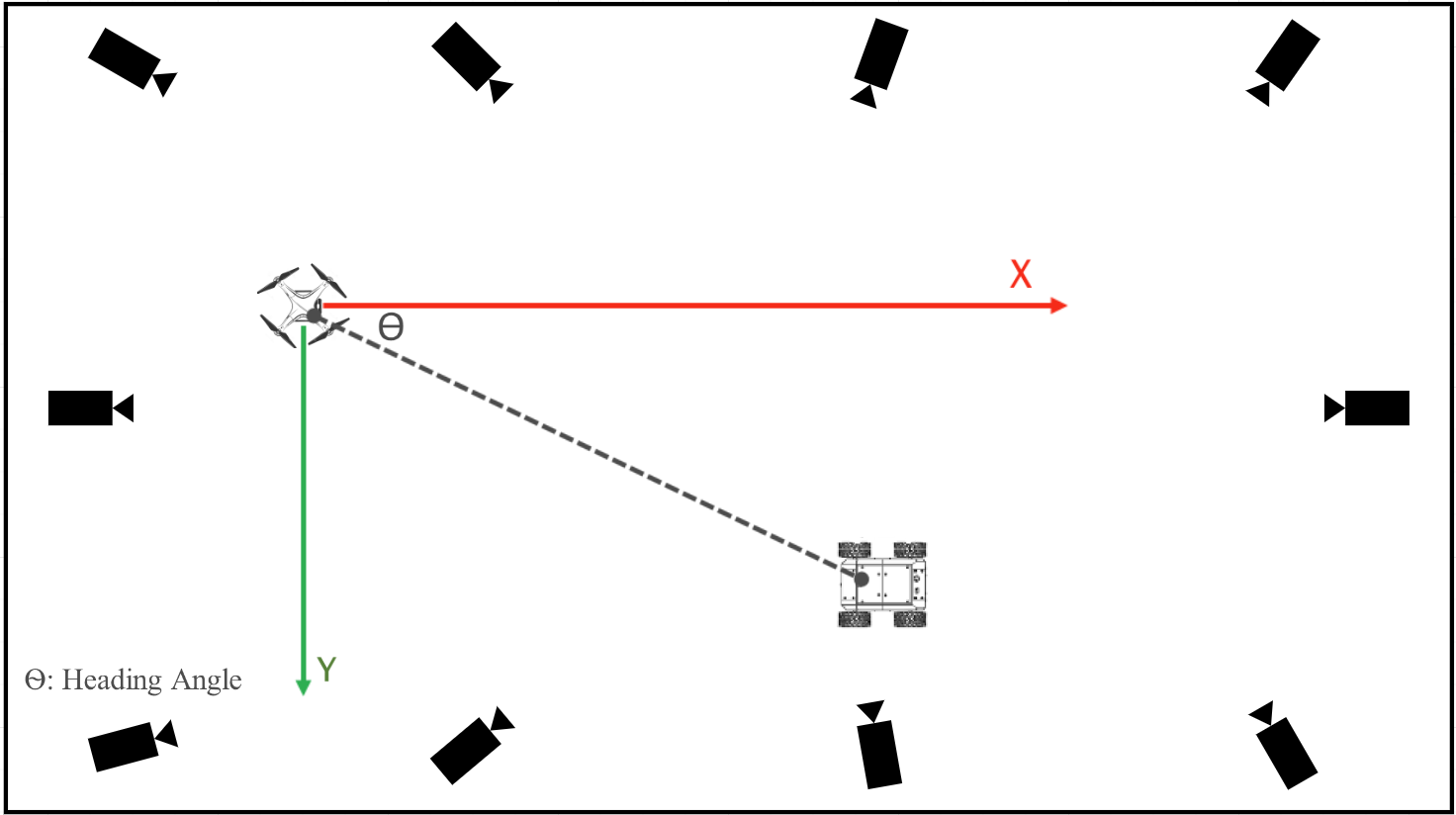}
    \caption{Top view of the laboratory scene showing VICON camera placement and UGV/UAV layout.}
    \label{fig:lab_topview}
\end{figure}

Data were collected in an indoor motion-capture arena equipped with a VICON system capable of full six degrees-of-freedom (6DoF) tracking for both the UAV and UGV, as illustrated in Figure~\ref{fig:lab_topview}. The VICON infrastructure provides millimeter-level positional accuracy, enabling high-fidelity relative pose annotation synchronized with image frames \cite{merriaux2017study}. During data collection, the UAV remained stationary while a single UGV traversed the scene under diverse headings and positions. This controlled setup was intentionally used to isolate the relationship between detector-derived bounding-box geometry and VICON-derived yaw correction without introducing additional compensation variables associated with moving-camera effects. Because the current framework operates on frame-level bounding-box descriptors, each image frame is treated as an independent observation of the relative target configuration in the UAV camera view \cite{ahmariconf,ahmarijournal}.

The UAV platform is equipped with two monocular cameras: a forward-facing camera (C1) used to observe the UGV for detection and heading estimation, and a downward-facing camera (C2) intended for landing verification once alignment is achieved. The present study uses only the forward-facing stream (C1), consistent with the goal of mapping image-plane bounding-box geometry to a yaw correction command.

The data collection pipeline integrates VICON tracking outputs with the ROS-based robotics stack using a hybrid Windows--Linux environment, as described in \cite{ahmariconf,ahmarijournal}. A custom Python interface is used to synchronize VICON logs and camera timestamps, producing frame-aligned records that associate each captured image with the corresponding ground-truth 6DoF state of the UAV and UGV. This synchronization is essential for producing reliable yaw supervision and for evaluating vision-based estimators under realistic detection noise and viewpoint variation.

Each image frame was annotated with a bounding box around the UGV to train a YOLO object detector tailored to the experimental environment \cite{ahmarijournal, ahmariconf}. As part of the detector-development procedure reported in a prior work, augmentation operations such as image flipping, rotation, brightness variation, and small translations were used to improve the robustness of UGV bounding-box detection. Once trained, the detector produces bounding boxes parameterized by pixel coordinates $(x_1,y_1,x_2,y_2)$ per frame. These coordinates form the basis for the low-dimensional geometric features used in this paper. While the original dataset contains additional metadata (including temporal alignment and 3D pose logs), the focus of the present work is on the mapping from bounding-box descriptors to yaw correction.

Ground-truth yaw labels are derived from VICON-based relative pose measurements as explained in equations \ref{eq:v1} and \ref{eq:v2}.


\section{Results and Discussion}\label{sec5}

This section reports an experimental evaluation of the proposed three-input fuzzy yaw controllers using five randomized train--test splits to quantify run-to-run variability. The evaluation is based on VICON-labeled image frames collected with a stationary UAV and a moving UGV; therefore, the reported results quantify yaw-estimation performance under controlled perception conditions. We first analyze the Mamdani controller as an interpretable baseline and then report the performance of the first-order Takagi--Sugeno controller with consequent parameters identified by least squares on the training split. All quantitative results are reported as mean $\pm$ standard deviation across runs.
\subsection{Mamdani Results}\label{subsec1}

The Mamdani fuzzy controller is designed as an interpretable baseline that promotes sign consistency between the lateral displacement of the target and the commanded yaw correction. Across five randomized train--test splits, the three-input Mamdani controller achieves a test-set MAE of $4.041^\circ \pm 0.054^\circ$, RMSE of $4.866^\circ \pm 0.052^\circ$, and MAXAE of $9.332^\circ \pm 0.091^\circ$. Within-threshold accuracies are $17.299\% \pm 1.658\%$, $41.005\% \pm 1.036\%$, and $64.279\% \pm 1.042\%$ for tolerance bands of $\pm1^\circ$, $\pm3^\circ$, and $\pm5^\circ$, respectively. These results indicate that, while the Mamdani controller reliably preserves the dominant turning direction, its fixed output membership functions and centroid defuzzification lead to limited numerical precision, particularly for small yaw corrections.

Beyond numerical accuracy, we evaluate SC between the image-plane lateral displacement and the predicted yaw sign using the same criterion across all runs. Averaged over five splits, the Mamdani controller achieves $91.183\% \pm 0.712\%$ directional agreement between the sign implied by $c_x$ and the sign of the predicted yaw. For completeness, Table~\ref{tab:sugeno_metrics} presents a comparative summary of the test-set performance of the Mamdani and Takagi--Sugeno models across all reported metrics.

\begin{figure}[h]
\centering
\includegraphics[width=0.6\textwidth]{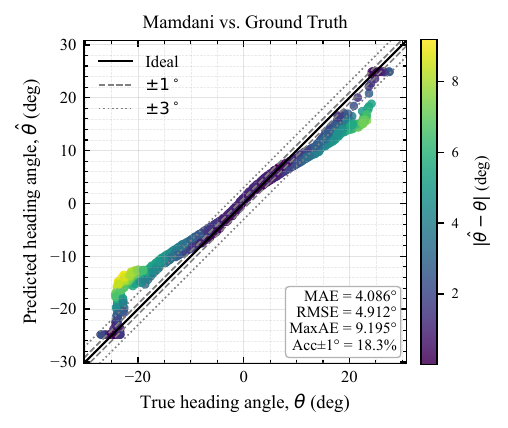}
\caption{Mamdani controller predictions versus ground-truth yaw on the test set. The solid black line denotes ideal prediction $\hat{\theta}=\theta$, while the dashed and dotted gray lines indicate $\pm1^\circ$ and $\pm3^\circ$ error bands, respectively. Points are colored by absolute prediction error $|\hat{\theta}-\theta|$. The representative split yields MAE $=4.086^\circ$, RMSE $=4.912^\circ$, MAXAE $=9.195^\circ$, and Acc$\pm1^\circ=18.3\%$.}
\label{fig:mamdani_scatter}
\end{figure}

Figure~\ref{fig:mamdani_scatter} shows the predicted yaw angle as a function of the ground-truth yaw on the test set. The predictions exhibit a clear monotonic relationship with the ground truth, indicating that the controller captures the dominant dependency between image-plane lateral displacement and yaw correction. Most predictions follow the identity-line trend, confirming that the same-side rule logic provides directionally meaningful behavior in the majority of cases.

However, structured deviations can be observed at larger yaw magnitudes, where the predicted responses bend away from the ideal line. This behavior is consistent with the fixed fuzzy output partitions used by Mamdani inference. These effects are inherent to the Mamdani formulation, which relies on fixed output membership functions and max--min inference.

\begin{figure}[h]
\centering
\includegraphics[width=0.6\textwidth]{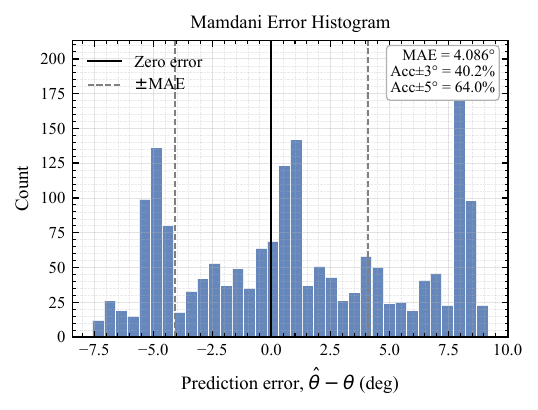}
\caption{Test-set error histogram for the Mamdani controller on a representative split. The solid black vertical line denotes zero error, and the dashed gray lines denote $\pm$MAE. The representative split yields MAE $=4.086^\circ$, Acc$\pm3^\circ=40.2\%$, and Acc$\pm5^\circ=64.0\%$.}
\label{fig:mamdani_error}
\end{figure}

The corresponding error histogram is shown in Figure~\ref{fig:mamdani_error}. The error distribution spans both negative and positive values and remains relatively broad, which is consistent with the moderate RMSE and the limited percentage of predictions within the tighter $\pm1^\circ$ and $\pm3^\circ$ bands.

This behavior is consistent with the limited expressiveness of fixed fuzzy consequents and the absence of data-driven tuning in the Mamdani framework.
Nevertheless, the Mamdani controller provides an interpretable baseline whose behavior is predictable and stable, making it suitable for safety-critical or explainability-focused applications. These limitations, however, motivate the adoption of a Takagi--Sugeno fuzzy inference system, which replaces fixed fuzzy consequents with data-driven local linear models to improve regression fidelity while retaining an interpretable rule structure.

\subsection{Takagi--Sugeno Results}\label{subsec2}

The first-order Takagi--Sugeno controller retains the same antecedent structure and rule base as the Mamdani controller but replaces fuzzy output sets with rule-local linear consequents identified via least-squares regression. This formulation preserves the interpretability of the fuzzy antecedents while substantially increasing expressive power for continuous regression.

Table~\ref{tab:sugeno_metrics} provides a consolidated comparison of the test-set performance of the Mamdani and Takagi--Sugeno controllers across all reported metrics, including numerical accuracy and directional side-consistency. Relative to the Mamdani baseline, the Takagi--Sugeno model achieves a marked reduction in both MAE and RMSE, along with a substantial improvement in within-threshold accuracy across all tolerance bands. Directional side-consistency remains high for both models. The Mamdani controller achieves slightly higher SC, consistent with its sign-oriented rule design, while the Takagi--Sugeno model remains comparably consistent despite learning its consequents from data.

\begin{table}[h]
\centering
\caption{Test-set performance comparison of Mamdani and Takagi--Sugeno controllers (3 inputs), reported as mean $\pm$ standard deviation over five randomized splits.}
\label{tab:sugeno_metrics}
\small
\setlength{\tabcolsep}{3.5pt}
\renewcommand{\arraystretch}{1.10}
\begin{tabular}{lccccccc}
\hline
Model & MAE & RMSE & MAXAE & $\pm1^\circ$ & $\pm3^\circ$ & $\pm5^\circ$ & SC \\
\hline
Mamdani 
& $4.04{\pm}0.05$ 
& $4.87{\pm}0.05$ 
& $9.33{\pm}0.09$ 
& $17.30{\pm}1.66$ 
& $41.01{\pm}1.04$ 
& $64.28{\pm}1.04$ 
& $\mathbf{91.18{\pm}0.71}$ \\

Sugeno  
& $\mathbf{0.14{\pm}0.00}$ 
& $\mathbf{0.20{\pm}0.01}$ 
& $\mathbf{1.25{\pm}0.12}$ 
& $\mathbf{99.68{\pm}0.27}$ 
& $\mathbf{100.00{\pm}0.00}$ 
& $\mathbf{100.00{\pm}0.00}$ 
& $90.25{\pm}0.61$ \\
\hline
\end{tabular}
\end{table}

\begin{figure}[h]
\centering
\includegraphics[width=0.55\textwidth]{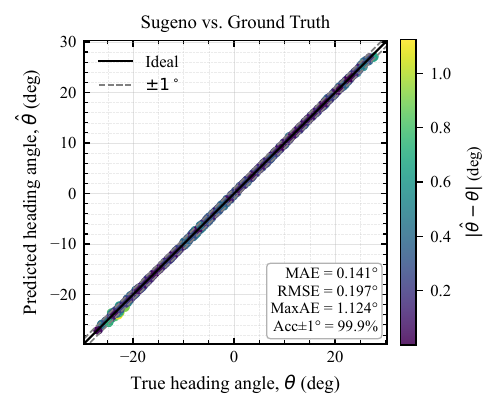}
\caption{Takagi--Sugeno controller predictions versus ground truth on the test set. The solid black line denotes ideal prediction $\hat{\theta}=\theta$, and the dashed gray lines indicate the $\pm1^\circ$ error band. Points are colored by absolute prediction error $|\hat{\theta}-\theta|$. The representative split yields MAE $=0.141^\circ$, RMSE $=0.197^\circ$, MAXAE $=1.124^\circ$, and Acc$\pm1^\circ=99.9\%$.}
\label{fig:sugeno_scatter}
\end{figure}

Figure~\ref{fig:sugeno_scatter} illustrates the predicted yaw versus ground truth on the test set.
In contrast to the Mamdani controller, the Sugeno predictions track the identity line more closely than the Mamdani controller across most of the yaw range. The point cloud remains tightly concentrated around the ideal line, and nearly all predictions fall within the $\pm1^\circ$ band, indicating that the learned rule-local linear consequents substantially improve numerical resolution without saturation.

Notably, the Sugeno controller maintains high accuracy near zero yaw while simultaneously preserving fidelity at large magnitudes, a regime where the Mamdani controller exhibits increased variance.

\begin{figure}[!h]
\centering
\includegraphics[width=0.55\textwidth]{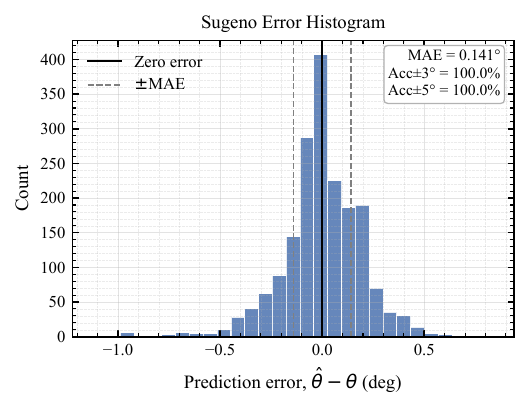}
\caption{Test-set error histogram for the Takagi--Sugeno controller. The solid black vertical line denotes zero error, and the dashed gray lines denote $\pm$MAE. The representative split yields MAE $=0.141^\circ$, Acc$\pm3^\circ=100.0\%$, and Acc$\pm5^\circ=100.0\%$.}
\label{fig:sugeno_error}
\end{figure}

The error histogram in Figure~\ref{fig:sugeno_error} is sharply peaked around zero,
with a near-zero mean error, indicating negligible systematic bias and substantially lower dispersion than the Mamdani baseline.
Across five randomized splits, the Takagi--Sugeno model achieves a test-set MAE of $0.140^\circ \pm 0.003^\circ$, RMSE of $0.200^\circ \pm 0.008^\circ$, and MAXAE of $1.254^\circ \pm 0.121^\circ$ (Table~\ref{tab:sugeno_metrics}). The model also achieves $99.676\% \pm 0.270\%$ accuracy within $\pm1^\circ$ and $100.000\% \pm 0.000\%$ accuracy within both $\pm3^\circ$ and $\pm5^\circ$.
Compared to the Mamdani controller, the Sugeno model substantially reduces the
frequency and magnitude of large errors and exhibits a more compact error distribution
around zero.

\begin{figure}[H]
\centering
\includegraphics[width=1\textwidth]{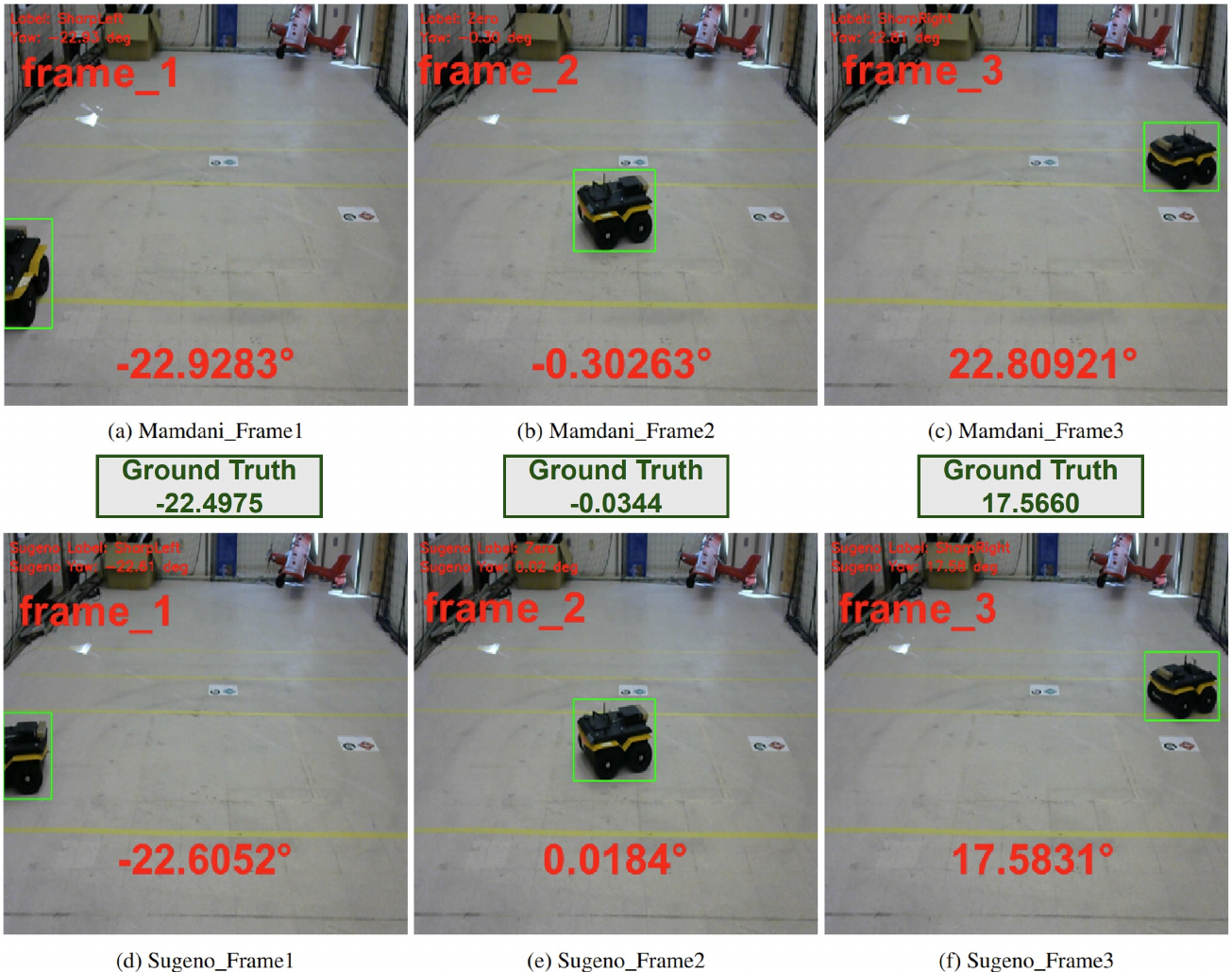}
\caption{Qualitative comparison on representative frames (left, center, and right yaw cases).
Each cell shows the same detected bounding box with overlayed yaw value: Predicted yaw $\hat{\theta}$ from the corresponding controller.
The Takagi--Sugeno model tracks the ground truth more closely across the full yaw range, while the Mamdani baseline exhibits larger deviations in magnitude under challenging cases.}
\label{fig:qualitative_frames}
\end{figure}

To complement the quantitative evaluation, Figure~\ref{fig:qualitative_frames} provides a frame-level qualitative comparison on representative test samples, showing the detected bounding boxes and the corresponding yaw predictions relative to the VICON ground truth.
The displayed frames are drawn from the held-out test set to represent left,
center, and right yaw cases using the same instances for both controllers,
ensuring a consistent qualitative comparison.

Table~\ref{tab:qualitative_diff} reports the absolute prediction error for the representative frames shown in Figure~\ref{fig:qualitative_frames}. While both controllers preserve the correct turning direction, the Mamdani controller exhibits larger magnitude errors, particularly in Frame~3, where output saturation effects become evident. In contrast, the Takagi–Sugeno model consistently achieves lower absolute error across all frames, including near-zero and large-magnitude yaw cases. This frame-level comparison reinforces the quantitative results and highlights the improved numerical fidelity of the Sugeno formulation.
\begin{table}[h]
\centering
\caption{Absolute prediction error corresponding to the qualitative examples shown in Figure~\ref{fig:qualitative_frames}.}
\label{tab:qualitative_diff}
\begin{tabular}{c c c}
\toprule
Frame & $|\Delta|$ Mamdani (deg) & $|\Delta|$ Sugeno (deg) \\
\midrule
Frame~1 & 0.43 & \textbf{0.11} \\
Frame~2 & 0.27 & \textbf{0.05} \\
Frame~3 & 5.25 & \textbf{0.02} \\
\bottomrule
\end{tabular}
\end{table}

Using the same side-consistency criterion adopted for the Mamdani evaluation, the Sugeno controller achieves $90.254\% \pm 0.612\%$ SC on the test set, indicating that the learned consequents largely preserve the intended same-side behavior while allowing deviations in visually ambiguous cases where minimizing regression error is favored.

The comparison between the Mamdani and Takagi--Sugeno controllers highlights a clear trade-off between interpretability and predictive accuracy.
The Mamdani controller offers transparent reasoning through fixed linguistic consequents and is designed to promote sign-consistent behavior through its same-side rule mapping, but its performance is limited by coarse output resolution and centroid defuzzification.
In contrast, the Sugeno controller preserves the same interpretable antecedent structure while significantly improving accuracy through least-squares identification of rule consequents.

Importantly, both controllers operate on the same three input features derived from YOLO bounding boxes.
Therefore, the observed performance improvement can be attributed to the inference and consequent formulation, rather than changes in feature representation.
These results demonstrate that the proposed Sugeno formulation provides a strong balance between interpretability and performance, making it well suited for vision-based yaw control in UAV--UGV interaction scenarios and motivating a broader comparison with standard low-dimensional regression baselines operating on the same features.

\section{Comparative Evaluation} 
\label{sec:comparative_evaluation} 
This section evaluates the proposed fuzzy inference models from three complementary perspectives. First, the models are compared with standard regression baselines operating on the same three bounding-box descriptors. Second, the computational volume and inference time of each method are reported. Third, the proposed yaw-estimation formulation is contrasted with image-based fuzzy visual servoing to clarify the difference between predicting an angular correction from bounding-box geometry and directly regulating image-plane error.

\subsection{Comparison with Standard Regression Baselines}
\label{subsec:baseline_comparison}

To evaluate the fuzzy inference models against conventional low-dimensional regression approaches, five standard regressors are considered using the same three bounding-box descriptors. The compared regressors include linear regression, ridge regression, support vector regression (SVR), random forest regression, and a shallow multilayer perceptron (MLP). All models are evaluated using the same five-run 70/30 train--test protocol and the same performance metrics used for the fuzzy models.

\begin{table}[h]
\centering
\caption{Comparison with standard regression baselines using the same three bounding-box features. Results are reported as mean $\pm$ standard deviation over five randomized train--test splits.}
\label{tab:baseline_comparison}
\scriptsize
\setlength{\tabcolsep}{3pt}
\renewcommand{\arraystretch}{1.10}
\begin{tabular}{lccccc}
\hline
Model & MAE ($^\circ$) & RMSE ($^\circ$) & MAXAE ($^\circ$) & Acc$\pm1^\circ$ (\%) & SC (\%) \\
\hline

Linear Regression 
& $0.668 \pm 0.009$ 
& $0.887 \pm 0.008$ 
& $5.443 \pm 0.242$ 
& $78.271 \pm 0.948$ 
& $90.124 \pm 0.555$ \\

Ridge Regression 
& $0.658 \pm 0.009$ 
& $0.889 \pm 0.010$ 
& $5.641 \pm 0.265$ 
& $79.082 \pm 0.895$ 
& $90.113 \pm 0.542$ \\

SVR 
& $0.135 \pm 0.003$
& $0.224 \pm 0.035$ 
& $3.045 \pm 3.359$ 
& $99.427 \pm 0.204$ 
& $90.189 \pm 0.549$ \\

Random Forest 
& $\mathbf{0.124 \pm 0.003}$ 
& $0.204 \pm 0.006$ 
& $1.404 \pm 0.181$ 
& $99.630 \pm 0.127$ 
& $90.232 \pm 0.553$ \\

Shallow MLP 
& $0.491 \pm 0.012$ 
& $0.632 \pm 0.025$ 
& $2.329 \pm 0.305$ 
& $87.618 \pm 2.556$ 
& $90.773 \pm 0.620$ \\

Mamdani 
& $4.041 \pm 0.054$ 
& $4.866 \pm 0.052$ 
& $9.332 \pm 0.091$ 
& $17.299 \pm 1.658$ 
& $\mathbf{91.183 \pm 0.712}$ \\

Takagi--Sugeno 
& $0.140 \pm 0.003$ 
& $\mathbf{0.200 \pm 0.008}$ 
& $\mathbf{1.254 \pm 0.121}$ 
& $\mathbf{99.676 \pm 0.270}$ 
& $90.254 \pm 0.612$ \\
\hline
\end{tabular}
\end{table}

The comparison shows that the three bounding-box descriptors contain strong predictive information for yaw estimation. Linear regression and ridge regression capture the dominant geometric trend, but their MAE, RMSE, and MAXAE remain substantially higher than those of the nonlinear regressors and the Takagi--Sugeno model. This indicates that a single global linear mapping is insufficient to fully represent the feature-to-yaw relationship in the present dataset.

Among the conventional regression baselines, random forest achieves the lowest average MAE, while SVR also provides competitive average error. However, the Takagi--Sugeno model achieves the lowest RMSE, lowest MAXAE, and highest Acc$\pm1^\circ$ among the evaluated models. The difference between random forest and Takagi--Sugeno in MAE is small, whereas the Takagi--Sugeno model provides a lower RMSE and a lower MAXAE, indicating a tighter overall and worst-case error profile under the reported test protocol. These results show that the Takagi--Sugeno formulation is competitive with standard low-dimensional regression baselines while preserving an interpretable fuzzy antecedent-rule structure. The Mamdani controller is less accurate as a continuous regressor, but it achieves the highest side-consistency, confirming its role as an interpretable directional baseline.
\subsection{Computational-Volume and Inference-Time Comparison}
\label{subsec:computational_volume}

For practical UAV applications, computational cost is an important consideration in addition to prediction accuracy. The fuzzy models and standard regression baselines are therefore compared in terms of general model structure, fitting time, and inference time per test sample. Timing values are measured on the same hardware and averaged over the test-set prediction procedure.

\begin{table}[h]
\centering
\caption{Computational-volume and inference-time comparison. Timing values are reported as mean $\pm$ standard deviation over five randomized train--test splits.}
\label{tab:computational_volume}
\scriptsize
\setlength{\tabcolsep}{3pt}
\renewcommand{\arraystretch}{1.10}
\begin{tabular}{lccc}
\hline
Model & Main structure & Fit time (s) & Inference time/sample (ms) \\
\hline
Linear Regression 
& Linear mapping 
& $0.001 \pm 0.001$ 
& $\mathbf{0.000053 \pm 0.000002}$ \\

Ridge Regression 
& Regularized linear mapping 
& $0.474 \pm 1.028$ 
& $0.000070 \pm 0.000018$ \\

SVR 
& RBF-kernel regression 
& $2.977 \pm 0.555$ 
& $0.084327 \pm 0.001127$ \\

Random Forest 
& Tree ensemble 
& $1.519 \pm 0.070$ 
& $0.013159 \pm 0.002836$ \\

Shallow MLP 
& One-hidden-layer neural regressor 
& $0.493 \pm 0.031$ 
& $0.000124 \pm 0.000016$ \\

Mamdani 
& Fuzzy rules + centroid defuzzification 
& $0.001 \pm 0.000$ 
& $0.090467 \pm 0.000571$ \\

Takagi--Sugeno 
& Fuzzy rules + weighted consequent average 
& $0.020 \pm 0.001$ 
& $0.000538 \pm 0.000020$ \\
\hline
\end{tabular}
\end{table}
The computational comparison shows that all evaluated methods operate on compact three-dimensional feature vectors rather than high-dimensional image inputs. Linear and ridge regression provide the lowest inference cost, but their prediction errors are substantially larger than those of the strongest nonlinear methods. The shallow MLP also has low inference time, but its numerical accuracy is lower than that of the Takagi--Sugeno model, SVR, and random forest. SVR provides competitive average error but has substantially higher measured inference time than the Takagi--Sugeno model. Random forest achieves the lowest MAE among the evaluated models, but it relies on an ensemble structure and has a higher measured inference time than the Takagi--Sugeno model.

The Takagi--Sugeno model provides a favorable compromise between accuracy, interpretability, and computational cost. Its inference is based on fuzzy rule activation, rule-local consequent evaluation, and normalized weighted averaging. Under the measured test protocol, it achieves substantially lower inference time than SVR, random forest, and Mamdani inference while maintaining competitive accuracy and the lowest reported MAXAE. The Mamdani model remains useful as an interpretable directional baseline, but its centroid-defuzzification step is computationally heavier in the present implementation.


\subsection{Comparison with Image-Based Fuzzy Visual Servoing}

A closely related vision-based fuzzy control framework is presented in \cite{visualServoFuzzyUAV}, where fuzzy inference is applied to reactive visual servoing of UAV yaw. In that approach, the controller inputs are the horizontal pixel residual
\[
e_x(k) = x_{\text{target}}(k) - x_{\text{center}},
\]
and its discrete-time derivative
\[
\Delta e_x(k) = e_x(k) - e_x(k-1),
\]
while the output is a yaw-related control command that drives the target toward image centering. The framework is validated through laboratory experiments and UAV flight tests, where performance is evaluated primarily using image-plane tracking errors and controller response trajectories.

Although both approaches employ fuzzy inference for vision-driven yaw behavior, they differ in representation, objective, and evaluation domain. The method in \cite{visualServoFuzzyUAV} operates directly in pixel coordinates. Under the standard pinhole projection model, pixel displacement depends on focal length and viewing geometry, implying that the magnitude of $e_x(k)$ varies with camera intrinsic parameters and target distance. As a result, membership function ranges and rule sensitivities are tied to the sensing configuration used during controller design, and transferring the controller across different camera setups may require recalibration.

The controller also incorporates the discrete derivative $\Delta e_x(k)$ to improve responsiveness. However, discrete differentiation can amplify high-frequency measurement noise, meaning that tracker jitter, partial occlusions, or bounding-box fluctuations may introduce transient spikes in the derivative term unless additional filtering is applied. This reflects a reactive visual servoing design in which instantaneous perception residuals are directly coupled to control actions.

The methodological differences between the image-based fuzzy servoing approach and the proposed framework are summarized in Table~\ref{tab:comparison}.

\begin{table}[h]
\centering
\small
\caption{Methodological comparison with the image-based fuzzy visual servoing framework in \cite{visualServoFuzzyUAV}.}
\label{tab:comparison}
\renewcommand{\arraystretch}{1.15}

\begin{tabularx}{\columnwidth}{p{2.9cm} p{5.1cm} p{4.1cm}}
\toprule
Aspect & Image-Based Fuzzy Servoing \cite{visualServoFuzzyUAV} & Proposed Framework \\
\midrule

Objective &
Image-plane yaw centering &
\textbf{Angular yaw estimation} \\

Inputs &
Pixel residual $e_x$ and derivative $\Delta e_x$ &
\textbf{Normalized geometric features} \\

Camera Dependence &
Residual magnitude tied to camera parameters &
Reduced pixel-scale dependence \\

Temporal Handling &
Discrete derivative term &
\textbf{No derivative input} \\

Noise Sensitivity &
Derivative amplifies tracking jitter &
Reduced sensitivity to measurement noise \\

Output &
Yaw control command &
\textbf{Yaw correction angle} \\

Evaluation &
Image-plane tracking error &
\textbf{Angular error (VICON ground truth)} \\

Inference Study &
Single fuzzy formulation &
\textbf{Mamdani vs.\ Sugeno} \\

\bottomrule
\end{tabularx}
\end{table}
The proposed framework instead formulates yaw alignment as an angular correction problem. Visual observations are mapped to resolution-normalized geometric features derived from detected bounding boxes, and fuzzy inference predicts a continuous yaw correction variable that is evaluated directly against independent ground-truth orientation measurements. By avoiding discrete error differentiation as a primary input, the proposed approach reduces sensitivity to derivative-induced noise amplification. Performance is therefore characterized statistically in the angular domain, enabling direct assessment of heading estimation accuracy under controlled experimental conditions.

\section{Conclusion}\label{sec6}

This paper introduced a structured and reproducible fuzzy-logic framework for estimating continuous yaw correction commands from compact vision-based descriptors derived from YOLO bounding boxes. By operating directly on low-dimensional geometric features, the proposed approach enables efficient and interpretable vision-to-control mapping without reliance on external localization at inference time or high-dimensional image representations.

Two fuzzy inference formulations were evaluated under identical conditions: a Mamdani controller serving as an interpretable baseline and a first-order Takagi--Sugeno model with consequents identified via least-squares regression. While the Mamdani system demonstrated reliable directional behavior and consistent sign alignment between lateral displacement and yaw output, its performance was constrained by the inherent discretization and saturation effects of fixed output membership functions. In contrast, the Takagi--Sugeno formulation preserved the same fuzzy antecedent structure while achieving substantially higher numerical accuracy, attaining a test-set MAE of $0.140^\circ \pm 0.003^\circ$, RMSE of $0.200^\circ \pm 0.008^\circ$, and MAXAE of $1.254^\circ \pm 0.121^\circ$ across five randomized splits, with $99.676\% \pm 0.270\%$, $100.000\% \pm 0.000\%$, and $100.000\% \pm 0.000\%$ of predictions within $\pm1^\circ$, $\pm3^\circ$, and $\pm5^\circ$, respectively. Directional agreement between image-plane lateral displacement and predicted yaw sign was maintained at $90.254\% \pm 0.612\%$, indicating that the learned consequents preserve the intended turning behavior in most cases.

A central contribution of this work lies in the controlled comparison of Mamdani and Takagi--Sugeno inference under a shared antecedent rule structure, training-derived preprocessing, and training-derived membership-function parameterization. This design clarifies how the consequent formulation affects vision-based yaw-estimation performance while preserving interpretability at the fuzzy-antecedent level. The comparative results show that the Mamdani controller remains useful for explainable directional reasoning, whereas the Takagi--Sugeno model provides substantially higher numerical precision and a favorable accuracy--interpretability--computational-cost trade-off relative to standard low-dimensional regression baselines. Importantly, all models operate on compact geometric features derived from detected bounding boxes and avoid reliance on high-dimensional image representations.
Overall, the proposed framework offers a practical, data-efficient, and computationally lightweight approach for vision-based UAV yaw guidance and establishes a clear methodological foundation for integrating fuzzy inference into perception-driven robotic control pipelines.

\section{Future Work}\label{sec7}
This work establishes a reproducible framework for mapping vision-derived
bounding-box geometry to continuous yaw commands using interpretable fuzzy
inference. Several extensions can further expand the capabilities of the
proposed approach.
Future work will evaluate the framework under dynamic UAV ego-motion and closed-loop flight conditions, where camera motion, detector jitter, and control latency may affect the observed bounding-box descriptors.

First, the current formulation operates on frame-level geometric descriptors.
Incorporating temporal information, such as short-term motion cues or
sequential feature aggregation, may improve robustness under rapid target
motion or intermittent detections. Second, the present design employs a
compact three-term fuzzy partition per input to preserve interpretability and
computational efficiency; future work may explore adaptive or hierarchical
fuzzy structures that allow richer partitions while maintaining transparent
rule representations.

Finally, extending evaluation beyond the controlled indoor VICON environment to more diverse operating conditions, including outdoor scenarios, illumination changes, background variability, different camera configurations, and alternative detector architectures, will provide further insight into the robustness of bounding-box-based fuzzy inference for perception-driven UAV guidance.

\bmhead{Acknowledgements}
This work was supported in part by the National Science Foundation (NSF) under Grant No. 2301553; in part by the National Aeronautics and Space Administration (NASA) University Leadership Initiative (ULI) under Grant Nos. 80NSSC20M0161 and 80NSSC25M7098; in part by the U.S. Department of Transportation (USDOT) under Grant No. 69A3552348327; and in part by the North Carolina Department of Transportation (NCDOT) under Grant No. RP2025-43.

\bibliography{ref}

@article{hutchinson2002tutorial,
  title={A tutorial on visual servo control},
  author={Hutchinson, Seth and Hager, Gregory D and Corke, Peter I},
  journal={IEEE transactions on robotics and automation},
  volume={12},
  number={5},
  pages={651--670},
  year={2002},
  publisher={IEEE}
}

@article{chaumette2006visual,
  title={Visual servo control. I. Basic approaches},
  author={Chaumette, Fran{\c{c}}ois and Hutchinson, Seth},
  journal={IEEE Robotics \& Automation Magazine},
  volume={13},
  number={4},
  pages={82--90},
  year={2006},
  publisher={IEEE}
}

@article{bojarski2016end,
  title={End to end learning for self-driving cars},
  author={Bojarski, Mariusz and Del Testa, Davide and Dworakowski, Daniel and Firner, Bernhard and Flepp, Beat and Goyal, Prasoon and Jackel, Lawrence D and Monfort, Mathew and Muller, Urs and Zhang, Jiakai and others},
  journal={arXiv preprint arXiv:1604.07316},
  year={2016}
}

@article{muhammad2020deep,
  title={Deep learning for safe autonomous driving: Current challenges and future directions},
  author={Muhammad, Khan and Ullah, Amin and Lloret, Jaime and Del Ser, Javier and De Albuquerque, Victor Hugo C},
  journal={IEEE Transactions on Intelligent Transportation Systems},
  volume={22},
  number={7},
  pages={4316--4336},
  year={2020},
  publisher={IEEE}
}

@article{mamdani1975experiment,
  title={An experiment in linguistic synthesis with a fuzzy logic controller},
  author={Mamdani, Ebrahim H and Assilian, Sedrak},
  journal={International journal of man-machine studies},
  volume={7},
  number={1},
  pages={1--13},
  year={1975},
  publisher={Elsevier}
}

@book{passino1998fuzzy,
  title={Fuzzy control},
  author={Passino, Kevin M and Yurkovich, Stephen and others},
  volume={42},
  year={1998},
  publisher={Addison-wesley Reading, MA}
}

@article{takagi1985fuzzy,
  title={Fuzzy identification of systems and its applications to modeling and control},
  author={Takagi, Tomohiro and Sugeno, Michio},
  journal={IEEE transactions on systems, man, and cybernetics},
  number={1},
  pages={116--132},
  year={1985},
  publisher={IEEE}
}

@book{sugeno1985industrial,
  title={Industrial applications of fuzzy control},
  author={Sugeno, Michio},
  year={1985},
  publisher={Elsevier Science Inc.}
}

@misc{hastie2009elements,
  title={The elements of statistical learning},
  author={Hastie, Trevor and Tibshirani, Robert and Friedman, Jerome and others},
  year={2009},
  publisher={Springer series in statistics New-York}
}

@article{loh2025theoretical,
  title={A theoretical review of modern robust statistics},
  author={Loh, Po-Ling},
  journal={Annual Review of Statistics and Its Application},
  volume={12},
  number={1},
  pages={477--496},
  year={2025},
  publisher={Annual Reviews}
}

@article{ahmarijournal,
  title={Visual Heading Prediction for Autonomous Aerial Vehicles},
  author={Ahmari, Reza and Mohammadi, Ahmad and Hemmati, Vahid and Mynuddin, Mohammed and Kebria, Parham and Mahmoud, Mahmoud Nabil and Yuan, Xiaohong and Homaifar, Abdollah},
  journal={arXiv preprint arXiv:2512.09898},
  year={2025}
}

@article{ahmariconf,
  title={A data-driven approach for uav-ugv integration},
  author={Ahmari, R and Hemmati, V and Mohammadi, A and Kebria, P and Mahmoud, M and Homaifar, A},
  journal={Proceedings of the Automation, Robotics \& Communications for Industry},
  volume={4},
  number={5.0},
  pages={77},
  year={2025}
}

@article{merriaux2017study,
  title={A study of vicon system positioning performance},
  author={Merriaux, Pierre and Dupuis, Yohan and Boutteau, R{\'e}mi and Vasseur, Pascal and Savatier, Xavier},
  journal={Sensors},
  volume={17},
  number={7},
  pages={1591},
  year={2017},
  publisher={MDPI}
}

@inproceedings{redmon2016you,
  title={You only look once: Unified, real-time object detection},
  author={Redmon, Joseph and Divvala, Santosh and Girshick, Ross and Farhadi, Ali},
  booktitle={Proceedings of the IEEE conference on computer vision and pattern recognition},
  pages={779--788},
  year={2016}
}

@article{chaumette2004image,
  title={Image moments: a general and useful set of features for visual servoing},
  author={Chaumette, Fran{\c{c}}ois},
  journal={IEEE Transactions on Robotics},
  volume={20},
  number={4},
  pages={713--723},
  year={2004},
  publisher={IEEE}
}

@article{wu2022survey,
  title={A survey of learning-based control of robotic visual servoing systems},
  author={Wu, Jinhui and Jin, Zhehao and Liu, Andong and Yu, Li and Yang, Fuwen},
  journal={Journal of the Franklin Institute},
  volume={359},
  number={1},
  pages={556--577},
  year={2022},
  publisher={Elsevier}
}

@article{kebria2019adaptive,
  title={Adaptive type-2 fuzzy neural-network control for teleoperation systems with delay and uncertainties},
  author={Kebria, Parham Mohsenzadeh and Khosravi, Abbas and Nahavandi, Saeid and Wu, Dongrui and Bello, Fernando},
  journal={IEEE Transactions on Fuzzy Systems},
  volume={28},
  number={10},
  pages={2543--2554},
  year={2019},
  publisher={IEEE}
}

@inproceedings{kebria2019fuzzy,
  title={Adaptive type-2 fuzzy control scheme for robust teleoperation under time-varying delay and uncertainties},
  author={Kebria, Parham M and Khosravi, Abbas and Jalali, Seyed Mohammad Jafar and Nahavandi, Saeid},
  booktitle={2019 IEEE 15th International Conference on Automation Science and Engineering (CASE)},
  pages={1631--1636},
  year={2019},
  organization={IEEE}
}

@inproceedings{kebria2019type,
  title={Type-2 fuzzy neural network synchronization of teleoperation systems with delay and uncertainties},
  author={Kebria, Parham M and Khosravi, Abbas and Jalali, Seyed Mohammad Jafar and Nahavandi, Saeid},
  booktitle={2019 IEEE 15th International Conference on Automation Science and Engineering (CASE)},
  pages={1625--1630},
  year={2019},
  organization={IEEE}
}

@article{liu2024explainable,
  title={Explainable attention-based UAV target detection for search and rescue scenarios},
  author={Liu, Shiyu and Yi, Ling and Xiong, Xuanrui and Tolba, Amr and Ding, Jinliang and Li, Chun},
  journal={IEEE Internet of Things Journal},
  year={2024},
  publisher={IEEE}
}

@article{ming2024not,
  title={Not all boxes are equal: Learning to optimize bounding boxes with discriminative distributions in optical remote sensing images},
  author={Ming, Qi and Miao, Lingjuan and Zhou, Zhiqiang and Vercheval, Nicolas and Pi{\v{z}}urica, Aleksandra},
  journal={IEEE Transactions on Geoscience and Remote Sensing},
  volume={62},
  pages={1--14},
  year={2024},
  publisher={IEEE}
}

@article{mendel2002type,
  title={Type-2 fuzzy sets made simple},
  author={Mendel, Jerry M and John, RI Bob},
  journal={IEEE Transactions on fuzzy systems},
  volume={10},
  number={2},
  pages={117--127},
  year={2002},
  publisher={IEEE}
}

@article{riza2015frbs,
  title={frbs: Fuzzy rule-based systems for classification and regression in R},
  author={Riza, Lala Septem and Bergmeir, Christoph and Herrera, Francisco and Ben{\'\i}tez, Jos{\'e} M},
  journal={Journal of statistical software},
  volume={65},
  pages={1--30},
  year={2015}
}

@article{visualServoFuzzyUAV,
  title={Visual servoing using fuzzy controllers on an unmanned aerial vehicles},
  author={Olivares M{\'e}ndez, Miguel {\'A}ngel and Campoy Cervera, Pascual and Mondragon Bernal, Ivan Fernando and Mart{\'\i}nez Luna, Carol Viviana},
  year={2009},
  publisher={Universidad P{\'u}blica de Navarra}
}

@article{karade2025robustyaw,
  title={Robust yaw angle control of autonomous underwater vehicle: dynamic surface-based optimized SoSMC},
  author={L. Karade, Vikas and V. Lakhekar, Girish and M. Shet, Raghavendra},
  journal={Intelligent Service Robotics},
  volume={18},
  number={4},
  pages={799--820},
  year={2025},
  publisher={Springer}
}

@article{mohsan2023unmanned,
  title={Unmanned aerial vehicles (UAVs): Practical aspects, applications, open challenges, security issues, and future trends},
  author={Mohsan, Syed Agha Hassnain and Othman, Nawaf Qasem Hamood and Li, Yanlong and Alsharif, Mohammed H and Khan, Muhammad Asghar},
  journal={Intelligent service robotics},
  volume={16},
  number={1},
  pages={109--137},
  year={2023},
  publisher={Springer}
}

\begin{appendices}

\section{Fuzzy Rule Base}\label{secA1}
This appendix lists the complete 27-rule antecedent grid used in both fuzzy systems.
The linguistic terms are defined as follows:
(i) $c_x$: \emph{Left}, \emph{Center}, \emph{Right};
(ii) $a$: \emph{Far}, \emph{Mid}, \emph{Near} (larger $a$ implies a closer target);
(iii) $r$: \emph{Wide}, \emph{Normal}, \emph{Tall} (larger $r$ implies a taller or narrower box).
The Mamdani consequent label $D_i\in\{\text{SharpLeft},\text{Left},\text{Zero},\text{Right},\text{SharpRight}\}$ is assigned by the same-side mapping implemented in our rule-generation function.
For the Takagi--Sugeno model, the antecedents are identical, but each rule uses a first-order consequent
$z_i(\mathbf{x})=p_i c_x + q_i a + s_i r + t_i$, where $(p_i,q_i,s_i,t_i)$ are learned by least squares on the training set.

\begin{table*}[!h]
\centering
\caption{Complete 27-rule base with requested linguistic terms. Sugeno uses the same antecedents with rule-local linear consequents $z_i(\mathbf{x})=p_i c_x + q_i a + s_i r + t_i$.}
\label{tab:rulebase27}
\scriptsize
\setlength{\tabcolsep}{4pt}
\begin{tabular}{c c c c c c}
\toprule
Rule $i$ & $c_x$ term & $a$ term & $r$ term & Mamdani consequent $D_i$ & Sugeno consequent $z_i(\mathbf{x})$ \\
\midrule
1  & Left   & Far  & Wide   & Left      & $p_1c_x+q_1a+s_1r+t_1$ \\
2  & Left   & Far  & Normal & Left      & $p_2c_x+q_2a+s_2r+t_2$ \\
3  & Left   & Far  & Tall   & SharpLeft & $p_3c_x+q_3a+s_3r+t_3$ \\
4  & Left   & Mid  & Wide   & Left      & $p_4c_x+q_4a+s_4r+t_4$ \\
5  & Left   & Mid  & Normal & Left      & $p_5c_x+q_5a+s_5r+t_5$ \\
6  & Left   & Mid  & Tall   & SharpLeft & $p_6c_x+q_6a+s_6r+t_6$ \\
7  & Left   & Near & Wide   & SharpLeft & $p_7c_x+q_7a+s_7r+t_7$ \\
8  & Left   & Near & Normal & SharpLeft & $p_8c_x+q_8a+s_8r+t_8$ \\
9  & Left   & Near & Tall   & SharpLeft & $p_9c_x+q_9a+s_9r+t_9$ \\
\midrule
10 & Center & Far  & Wide   & Zero      & $p_{10}c_x+q_{10}a+s_{10}r+t_{10}$ \\
11 & Center & Far  & Normal & Zero      & $p_{11}c_x+q_{11}a+s_{11}r+t_{11}$ \\
12 & Center & Far  & Tall   & Zero      & $p_{12}c_x+q_{12}a+s_{12}r+t_{12}$ \\
13 & Center & Mid  & Wide   & Zero      & $p_{13}c_x+q_{13}a+s_{13}r+t_{13}$ \\
14 & Center & Mid  & Normal & Zero      & $p_{14}c_x+q_{14}a+s_{14}r+t_{14}$ \\
15 & Center & Mid  & Tall   & Zero      & $p_{15}c_x+q_{15}a+s_{15}r+t_{15}$ \\
16 & Center & Near & Wide   & Zero      & $p_{16}c_x+q_{16}a+s_{16}r+t_{16}$ \\
17 & Center & Near & Normal & Zero      & $p_{17}c_x+q_{17}a+s_{17}r+t_{17}$ \\
18 & Center & Near & Tall   & Zero      & $p_{18}c_x+q_{18}a+s_{18}r+t_{18}$ \\
\midrule
19 & Right  & Far  & Wide   & Right     & $p_{19}c_x+q_{19}a+s_{19}r+t_{19}$ \\
20 & Right  & Far  & Normal & Right     & $p_{20}c_x+q_{20}a+s_{20}r+t_{20}$ \\
21 & Right  & Far  & Tall   & SharpRight& $p_{21}c_x+q_{21}a+s_{21}r+t_{21}$ \\
22 & Right  & Mid  & Wide   & Right     & $p_{22}c_x+q_{22}a+s_{22}r+t_{22}$ \\
23 & Right  & Mid  & Normal & Right     & $p_{23}c_x+q_{23}a+s_{23}r+t_{23}$ \\
24 & Right  & Mid  & Tall   & SharpRight& $p_{24}c_x+q_{24}a+s_{24}r+t_{24}$ \\
25 & Right  & Near & Wide   & SharpRight& $p_{25}c_x+q_{25}a+s_{25}r+t_{25}$ \\
26 & Right  & Near & Normal & SharpRight& $p_{26}c_x+q_{26}a+s_{26}r+t_{26}$ \\
27 & Right  & Near & Tall   & SharpRight& $p_{27}c_x+q_{27}a+s_{27}r+t_{27}$ \\
\bottomrule
\end{tabular}
\end{table*}




\end{appendices}


\end{document}